\documentclass[twocolumn]{cinc}
\usepackage{mathtools, graphicx, xurl, hyperref}
\usepackage{amsfonts}
\usepackage{multirow}
\usepackage{xcolor}

\title{A Mixed-Domain Self-Attention Network for Multilabel Cardiac Irregularity Classification Using Reduced-Lead Electrocardiogram}
\author{Hao-Chun Yang\textsuperscript{1*},
Wan-Ting Hsieh\textsuperscript{2*},
Trista Pei-Chun Chen\textsuperscript{2} \\ \ \\
\textsuperscript{1}National Tsing Hua University, Hsinchu, Taiwan \\
\textsuperscript{2}Inventec Corporation, Taipei, Taiwan
\thanks{* The two authors contributed equally to this paper.}}

\begin{document}
\maketitle
\begin{abstract}
Electrocardiogram(ECG) is commonly used to detect cardiac irregularities such as atrial fibrillation, bradycardia, and other irregular complexes. While previous studies had great success classifying these irregularities with standard 12-lead ECGs, there existed limited evidence demonstrating the utility of reduced-lead ECGs in capturing a wide-range of diagnostic information. In addition, classification model's generalizability across multiple recording sources also remained uncovered. As part of the PhysioNet/Computing in Cardiology Challenge 2021, our team HaoWan\_AIeC, proposed \textbf{Mixed-Domain Self-Attention Resnet (MDARsn)} to identify cardiac abnormalities from reduced-lead ECG. Our classifiers received scores of 0.4, 0.33, 0.37, 0.34, and 0.34 (ranked 18th, 23rd, 20th, 23rd, and 22nd) for the 12-lead, 6-lead, 4-lead, 3-lead, and 2-lead versions of the hidden test set with the evaluation metric defined by the challenge.

\end{abstract}
\section{Introduction}
Cardiovascular diseases (CVDs) can be life-threatening which causes 17.9 million deaths each year. Early diagnosis of cardiac abnormalities is crucial as it can prevent complications and improve treatment outcomes\cite{vitale2004cardiac}. The standard 12-lead electrocardiogram (ECG) is widely used to monitor cardiac function. However, the accessibility to 12-lead ECG is limited. The PhysioNet/Computing in Cardiology Challenge 2021 focused on automated, open-source approaches to classify cardiac abnormalities from the reduced set of leads (reduced-lead ECGs) \cite{2020ChallengePMEA, 2021ChallengeCinC}.
With reduced-lead ECGs, it is noted that signals from different leads are helpful in various CVDs diagnosis\cite{gunnarsson2001ecg}. Methods that automatically learn the relationship between ECG leads and CVDs are desired.

To tackle domain discrepancy between the training and test sets, the work \cite{hasani2020classification} learned domain-invariant features with  domain-adversarial training by perturbing signals with adversarial gradients and augmenting model-based data. The result however was unsatisfactory when the test sets contain unseen data during training.

\begin{figure}[tpb]
\centering
\scalebox{.35}{\includegraphics{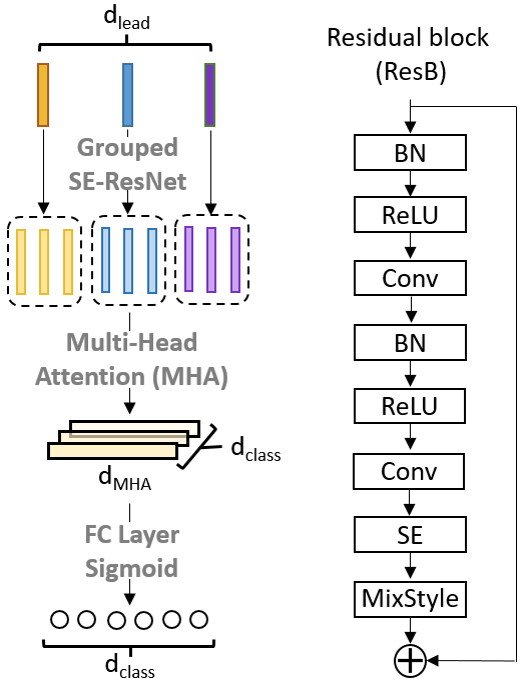}}
\caption{Left: the proposed architecture. Right: the residual block(ResB) with MixStyle blocks.}
\label{fig:label}
\end{figure}


\section{Methods}

In this work, we present a Mixed-Domain self-Attention ResNet (MDARsn) to achieve two goals: coping with domain bias among different sources or between training and hidden datasets, and learning the relationships between ECG leads and CVDs automatically. We adopt feature-based augmentation method from \cite{zhou2021domain} and Multi-Head Attention (MHA) layer to reach two objectives.


\subsection{Data Pre-processing}\label{sec: dataprep}
\textbf{Resampling} 
Public training data provided from 2021 PhysioNet / CinC challenge were sourced from $5$ locations. As recordings from separate hospitals could have different sampling rates, we upsampled or downsampled each recording to 500 Hz. Each recording was filtered by an FIR (finite impulse response) bandpass filter with bandwidth between $3-45$ Hz. Later we applied min-max normalization to every recording for the signal to range between $-1$ and $1$.

\begin{figure}[tpb]
\centering
\scalebox{.25}{\includegraphics{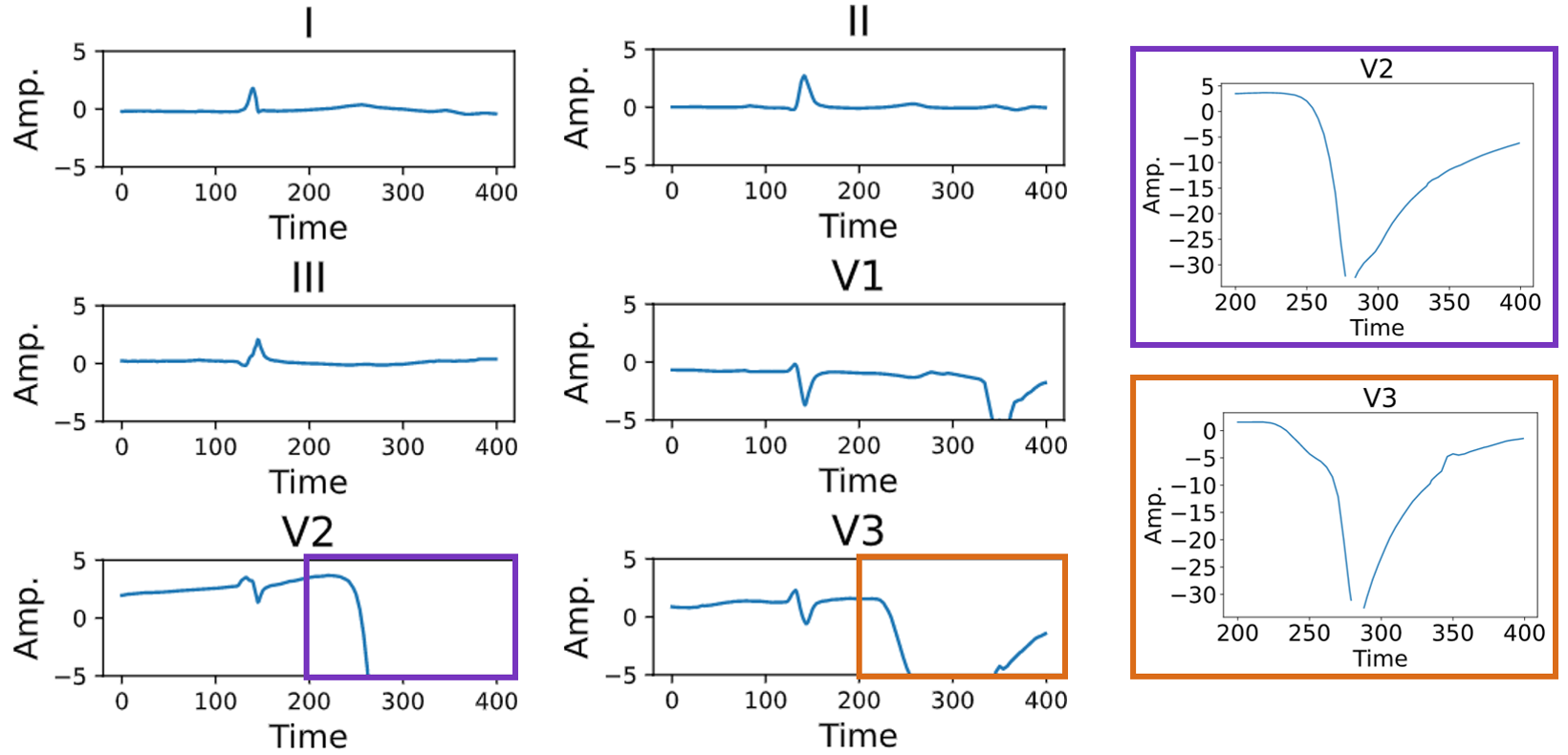}}
\caption{The chunked ECG signal of WFDB\_Ga/E06072. The \textit{NaN} fragments are found in lead V2 and V3, along with rapidly decreasing and increasing trend before and after the \textit{NaN} fragment.}
\label{fig:invalid}
\end{figure}

\textbf{Valid lead checking} 
We observed that some leads contain non-numerical data (\textit{NaN}).
To avoid weighting on low-quality leads, we created a valid-lead mask for each recording where the ECG leads with \textit{NaN} are marked as $0$ and the rest are marked as $1$. Such valid-lead mask is later passed to the MHA layer to avoid weighting on low-quality leads.


\textbf{Data augmentation} 
For better training, we augmented data by randomly cropping the signal and randomly generating valid-lead mask. Given a recording, we extracted a random window size of data. If the sequence is less than $T=15$ seconds, we padded the sequence with zeros. The valid-lead mask for the sequence is randomly assigned with $0$s, to simulate the broken lead situation seen in the real-world.
\color{black}

\subsection{Model Architecture}\label{sec: modelArch}


\subsubsection{Squeeze-and-Excitation ResNet (SERsn)}
SERsn is reported a powerful framework to model ECG signals for CVD detection \cite{zhao2020adaptive}. The proposed SERsn consisted of one convolution layer followed by $N=8$ residual blocks (ResBs), each of which contained two convolution layers and a squeeze-and-excitation (SE) block \cite{hu2018squeeze}. The number of filters increased by a factor of two for every two ResBs. We grid-searched on the kernel size in $[7,9,11,13,15,17]$ and found similar conclusion as \cite{zhao2020adaptive} that larger kernel size at the first convolution layer and smaller kernel size for the rest achieved better results. We further modified the sequence of layers, applying batch-normalization first, followed by ReLU activation and then convolution layer. 

\subsubsection{Domain Generalization}

Since the training datasets are collected from different sources, and there exists data discrepancy between the training and the hidden sets, we adopted MixStyle\cite{zhou2021domain} which has demonstrated its generalization capability with feature-based augmentation.

MixStyle is derived from instance normalization which could effectively remove instance-specific characteristic, while it further mixes the feature statistics of feature instances from two domains with a random weight. The basic implementation of instance normalization is as defined in Equation \ref{eq:in}, which normalizes feature maps ${F}$ with means and standard deviations computed within each channel, followed by the affine transformation. 
\begin{equation}
IN(F) = \gamma \odot \frac{F-\mu(F)}{\sigma(F)} + \beta
\label{eq:in}
\end{equation}
Where ${F}\in$$\mathbb{R}$$^{B\times C\times H\times W}$ with $B$, $C$, $H$ and
$W$ denoting batchsize, channel, height and width respectively. $\gamma,\beta \in $$\mathbb{R}$$^{C}$ are the affine transformation coefficient. Specifically, for $c \in {1, ..., C}$,
\begin{equation}
\mu(F)_c = \frac{1}{BHW}\sum_{b=1}^B \sum_{h=1}^H \sum_{w=1}^W F_{b,c,h,w},
\label{eq:mean_in}
\end{equation}
and
\begin{equation}
\sigma(F)_c = \sqrt{\frac{1}{BHW} \sum_{b=1}^B \sum_{h=1}^H \sum_{w=1}^W(F_{b,c,h,w} - \mu(F)_c)^2}
\label{eq:std_in}
\end{equation}
Given two sets of feature instances $F$ and $F'$, MixStyle performs domain generalization by generating a mixture of feature statistics as defined in Equation \ref{eq:ms}, 
\begin{equation}
MixStyle(F, F') = \gamma_{mix}\odot \frac{F-\mu(F)}{\sigma(F)} + \beta_{mix}
\label{eq:ms}
\end{equation}

\begin{equation}
\gamma_{mix} = \lambda\sigma(F) + (1-\lambda)\sigma(F'), 
\label{eq:gamma}
\end{equation}

\begin{equation}
\beta_{mix} = \lambda\mu(F) + (1-\lambda)\mu(F')
\label{eq:beta}
\end{equation}
where $\lambda$ is random weight sampled from beta distribution, $\lambda \sim Beta(\alpha)$, with $\alpha=0.1$. As $F$ and $F'$ requires two sources, the simple solution is to shuffle the order of the batch dimension of $F$ to obtain $F'$, which is validated to attain performance comparable to cross-domain mixing\cite{zhou2021domain}. 
MixStyle blocks are inserted after SE-block in ResBs, besides they are only applied to shallow ResBs blocks which capture more domain-related information. We tuned the number of layers that applied MixStyle, ${n}_{mix}$, and found adding it to previous $1\sim2$ ResBs is enough.



\subsubsection{Mixed-Domain self-Attention SERsn (MDARsn)}
The proposed MDARsn aims to learn the domain-invariant and lead-independent representation. It ignores readings from broken leads and automatically weights on the rest of the leads for final prediction. The proposed MDARsn has three modules: (1) grouped SERsn to learn the lead-independent representation; (2) feature-based data augmentation, MixStyle\cite{zhou2021domain}, to generalize for different domains; (3) MHA layer to mask low quality ECG leads and to weight on the rest lead-independent representation.

We learn the lead embedding from SERsn by setting the cardinality to the number of leads. Under this setting, the input channels would be divided in groups (i.e, each lead forms a group) and would learn different types of features while increasing the efficiency of weights\cite{xie2017aggregated} through this groups SERsn. In addition, the MixStyle blocks are added to shallow ResBs. Once the lead embedding is extracted from SERsn, we use the MHA layer to jointly attend to information from different ECG leads\cite{vaswani2017attention}.
The MHA is based on dot-product attention as defined in below:
\begin{equation}
Attention(Q,K,V) = softmax(\frac{Q{K}^{T}}{\sqrt{d_k}})
\label{eq:mha}
\end{equation}
Where Q, K, V are query, key and value vector derived by linear projection of lead embedding, while $d_k$ denotes the dimension of key vector.
We set the number of heads as the number of classes, ${d}_{class}$, and choose the embedding dimension ${d}_{model}=520$. We input the valid-lead mask described in section \ref{sec: dataprep} to MHA so that only the embedding of valid leads would be considered in the final classification. Ultimately, for every recording we obtain an output from MHA layer with dimension (${d}_{class}$, ${d}_{model}/{d}_{class}$), which would be later passed into a fully connected layer and output (${d}_{class}$, $1$).

\subsection{Implementation Details}
To reduce the training and validation time, we excluded samples that were not labeled in one of the $26$ classes that will be considered for challenge metric\cite{2021ChallengeCinC}. Besides, we also excluded recordings from PTB \& StPetersburg as described in \cite{9344302} due to the long average lengths. We then split the samples from the rest of datasets\cite{CPSC, PTB-XL, Chapman-Shaoxing, Ningbo} into $5$-fold using iterative stratification to guarantee the label balance in each folds\cite{pmlr-v74}. We conducted all experiments using $3$ folds as training, $1$ fold as validation and $1$ fold to test. During model training we monitored validation precision score (PRC) and used early stopping when validation PRC had stopped improving for $5$ epochs.

We utilized a standard binary cross entropy loss function averaged over $26$ classes to train the model. The optimizer is Adam and a cosine annealing was applied with warm-up $1000$ steps and maximum $10000$ steps. Several hyperparameters were greedily searched using Optuna Toolkit \cite{optuna_2019} monitoring on cross-validated testing set's challenges metric: learning rate is searched within $[0.0001\sim0.001]$, dropout $[0.1\sim0.5]$, ${n}_{mix}$ among $[1\sim3]$, convolution kernel sizes $[5,7,9,11,13,15,17]$ and number of stacked ResBs $[6,8,10,12]$. Models were trained using PyTorch on two 2080-Ti GPUs with a batch size of $96$. Each epoch took roughly $5$ minutes to train and around $2.5$ hours to complete a parameter set. The final best searched parameters were listed in Table \ref{tab:params}.


\begin{table}[t]
    \scalebox{0.8}{%
    \begin{tabular}{lccccc}
        \hline
        Leads                                  & 12  & 6   & 4   & 3   & 2   \\ \hline
        General                                &     &     &     &     &     \\ \hline
        ECG window size (secs)                 & 15  & 15  & 15  & 15  & 15  \\
        Sampling   frequency (Hz)              & 500 & 500 & 500 & 500 & 500 \\
        Number of classes, ${d}_{class}$              & 26  & 26  & 26  & 26  & 26  \\ \hline
        SERsn                             &     &     &     &     &     \\ \hline
        Input channel                          & 12  & 6   & 4   & 3   & 2   \\
        Channel of the first conv. layer       & 128 & 128 & 128 & 128 & 128 \\
        kernel   size of the first conv. layer & 11  & 11  & 11  & 11  & 11  \\
        Kernel size of ResBs                   & 7   & 5   & 13  & 7   & 7   \\
        Stride                                 & 3   & 3   & 3   & 3   & 3   \\ \hline
        MixStyle                                    &     &     &     &     &     \\ \hline
        Number of applied layers, ${n}_{mix}$                                & 2    & 2    & 2    & 2    & 1  \\ \hline
        MHA                                    &     &     &     &     &     \\ \hline
        Embedding   size, ${d}_{model}$               & 650 & 520 & 650 & 520 & 650 \\
        Number of heads, ${d}_{head}$                 & 26  & 26  & 26  & 26  & 26  \\ \hline
        \end{tabular}}
\caption{Hyperparamters searched for different leads.}
\label{tab:params}
\end{table}

\section{Results}
\begin{table}[t]
    \scalebox{0.8}{%
        \begin{tabular}{llllll}
        \hline
        Model                                                                               & Leads & Training       & Validation     & Test & Ranking \\ \hline
        \multirow{5}{*}{SERsn}                                                          & 12    & 0.721          & 0.582          & -    & -       \\
                                                                                            & 6     & 0.7            & 0.576          & -    & -       \\
                                                                                            & 4     & 0.704          & 0.580          & -    & -       \\
                                                                                            & 3     & 0.704          & 0.586          & -    & -       \\
                                                                                            & 2     & 0.699          & 0.580          & -    & -       \\ \hline
        \multirow{5}{*}{SERsn+M} & 12    & 0.731          & \textbf{0.602} & 0.4  & 18     \\
                                                                                            & 6     & 0.709          & \textbf{0.593} & 0.33  & 23     \\
                                                                                            & 4     & 0.711          & \textbf{0.597} & 0.37  & 20     \\
                                                                        \textbf{}                    & 3     & 0.713          & \textbf{0.591} & 0.34  & 23     \\
                                                                                            & 2     & 0.705          & \textbf{0.589} & 0.34  & 22     \\ \hline
        \multirow{5}{*}{\begin{tabular}[c]{@{}l@{}}SERsn+M+A \\ (\textbf{MDARsn})\end{tabular}}                                                          & 12    & \textbf{0.738} & 0.525          & -    & -       \\
                                                                                            & 6     & \textbf{0.71}  & 0.506          & -    & -       \\
                                                                                            & 4     & \textbf{0.723} & 0.511          & -    & -       \\
                                                                                            & 3     & \textbf{0.719} & 0.503          & -    & -       \\
                                                                                            & 2     & \textbf{0.707} & 0.499          & -    & -       \\ \hline
        \end{tabular}}
\caption{Challenge scores for different models testing on training set, repeated scoring on the hidden validation set, and one-time scoring on the hidden test set as well as the ranking on the hidden test set. M: with MixStyle block; A: with MHA layer.}
\label{tab:results}
\end{table}

Table \ref{tab:results} reports the Challenge score \cite{2021ChallengeCinC} on the training, hidden validation and test sets of three models: SERsn, SERsn with mixStyle (SERsn+M) and the proposed MDARsn. We demonstrate that with the MixStyle blocks, the results outperform the plain SERsn in both training and hidden validation sets. We also obtain better performance using Attn-Rsn in the training set, increasing at least $0.02$ of the Challenge score. However, there is a drop with the proposed MDARsn of approximately $0.08$ in the hidden validation set comparing to SERsn+M. As SERsn with mixStyle shows the steady performance overall, we choose it as our final model and report its results on hidden test set in Table \ref{tab:results}.

\section{Discussion and Conclusions}
In this study, we proposed Mixed-Domain self-Attention ResNet (MDARsn) to classify $26$ CVDs from reduced-lead ECGs. Domain-specific characteristics are removed by the \textit{MixStyle} block. With its source invariant representation, the classifier is robust when encountering unseen ECG recordings. We also incorporated the \textit{Multi-Head Self-Attention} block to handle missing or low-quality reduced-lead ECGs. Both measures provided improved classification on the local nested training set.


Our model exhibited two weaknesses. First we found a performance gap between the training set and the hidden validation/test set provided by the challenge host. Second, we found it under-performing on the unseen local datasets (PTB \& StPetersburg). These two weaknesses suggested that we further investigate our data preprocessing pipeline for different recording sources as MHA did not seem to handle different data sources as well as expected. We hope to further improve MDARsn's generalizability and flexibility in reduced-lead ECG classification.

\color{black}



\bibliographystyle{cinc}
\bibliography{cinc_template} 

\vspace{-5mm}
\begin{correspondence}
Wan-Ting Hsieh\\
111 No. 166, Sec. 4, Chengde Rd., Shilin Dist., Taipei, Taiwan\\
hsieh.eileen@inventec.com
\end{correspondence}


\end{document}